\documentclass[10pt,twocolumn,letterpaper]{article}

\usepackage{cvpr}
\usepackage{times}
\usepackage{epsfig}
\usepackage{graphicx}
\usepackage{amsmath}
\usepackage{amssymb}
\usepackage{multirow}
\usepackage{enumitem}
\usepackage{booktabs}
\usepackage{tabularx}
\usepackage{xcolor}

\def\eg{\textit{e.g. }}
\def\ie{\textit{i.e.}}

\def\etal{\textit{et al.}}

\newcommand{\vect}[1]{\mathbf{#1}}

\usepackage[breaklinks=true,bookmarks=false]{hyperref}

\cvprfinalcopy 

\def\cvprPaperID{4818} 

\ifcvprfinal\fi
\begin{document}

\title{Refine and Distill: Exploiting Cycle-Inconsistency and Knowledge \\ Distillation for Unsupervised Monocular Depth Estimation}

\author{Andrea~Pilzer$^1$,
        ~St\'{e}phane~Lathuili\`{e}re$^1$,
        ~Nicu~Sebe$^{1, 2}$,
        and~Elisa~Ricci$^{1, 3}$\\
	$^1$DISI, University of Trento, via Sommarive 14, Povo (TN), Italy\\
	$^2$Huawei Technologies Ireland, Dublin, Ireland\\
	$^3$Technologies of Vision, Fondazione Bruno Kessler, via Sommarive 18, Povo (TN), Italy\\
{\tt\small \{andrea.pilzer, stephane.lathuiliere, niculae.sebe, e.ricci\}@unitn.it}
}

\maketitle

\begin{abstract}
Nowadays, the majority of state of the art monocular depth estimation techniques are based on supervised deep learning models. 
However, collecting RGB images with associated depth maps is a very time consuming procedure. Therefore, recent works have proposed deep architectures for addressing the monocular depth prediction task as a reconstruction problem, thus avoiding the need of collecting ground-truth depth. Following these works, we propose a novel self-supervised deep model for estimating depth maps. Our framework exploits two main strategies: refinement via cycle-inconsistency and distillation. Specifically, first a \emph{student} network is trained to predict a disparity map such as to recover from a frame in a camera view the associated image in the opposite view. Then, a backward cycle network is applied to the generated image to re-synthesize back the input image, estimating the opposite disparity. A third network exploits the inconsistency between the original and the reconstructed input frame in order to output a refined depth map. Finally, knowledge distillation is exploited, such as to transfer information from the refinement network to the student. Our extensive experimental evaluation demonstrate the effectiveness of the proposed framework which outperforms state of the art unsupervised methods on the KITTI benchmark. 
\end{abstract}

\section{Introduction}

In the last few years, deep learning-based approaches for 
depth estimation ~\cite{eigen2014depth,ladicky2014pulling,liu2016learning,xu2018monocular,godard2017unsupervised,zhou2017unsupervised,mahjourian2018unsupervised,pilzer2018unsupervised} have attracted a growing interest, motivated, on the one hand, by their ability to predict very accurate depth maps and, on the other hand, by the importance of recovering depth information in several applications, such as robot navigation, autonomous driving, virtual reality and 3D reconstruction.

\begin{figure}[!t]
\begin{center}
\includegraphics[width=0.8\columnwidth]{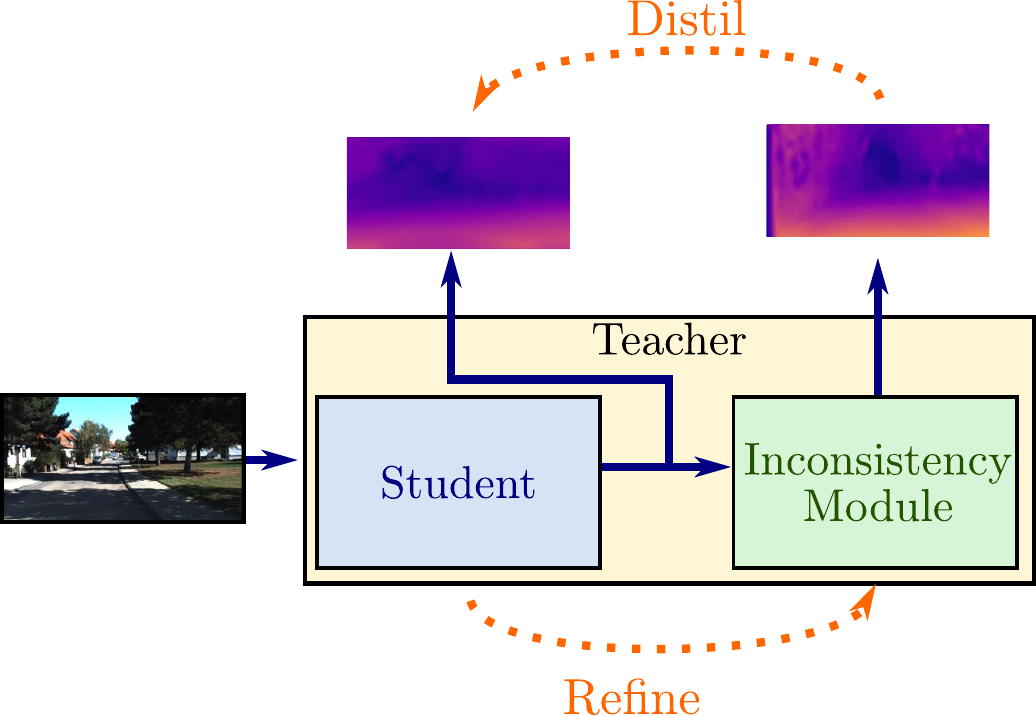}
\caption{Outline of the proposed approach: from the right view image, we predict the left image from which we re-synthesize the right image. The inconsistencies are used by the inconsistency-module to improve the depth estimation. The refined depth maps are used to improve the Student Network via knowledge distillation.}
\label{fig:teaser}
\end{center}
\vspace{-25pt}
\end{figure}

Exploiting the availability of very large annotated datasets, Convolutional Neural Networks (ConvNets) trained in a supervised setting are now state-of-the-art in many computer vision tasks such as object detection~\cite{girshick2015fast}, instance segmentation~\cite{Novotny_2018_ECCV}, human pose estimation~\cite{Nie_2018_ECCV}. However, a major weakness of these approaches is the need of collecting large-scale labeled datasets. 
In the case of depth estimation, acquiring data is especially costly. For instance, in the scenario of depth estimation for autonomous driving, it implies driving a car equipped with a laser LiDaR scanner for hours under diverse lighting and weather conditions.
Self-supervised depth estimation, also referred to as unsupervised, recently emerged as an interesting paradigm and an effective alternative to supervised methods~\cite{luo2016efficient,garg2016unsupervised,mayer2016large,godard2017unsupervised,wang2017learning}. Roughly speaking, in the self-supervised setting, stereo image pairs are considered as input and a deep predictor is learned in order to estimate the associated disparity maps. 
Specifically, the predicted disparity is employed to synthesize, from a frame in a camera view (\textit{e.g.} from the left camera), the opposite view through warping. 
The deep network is trained via gradient descent by minimizing the discrepancy between the original and the reconstructed image. 
Importantly, even if stereo images pairs are required for training, depth can be recovered from a single image at test time.

In this paper, we follow this research thread and propose a novel self-supervised deep architecture for monocular depth estimation. The proposed approach, illustrated in Fig \ref{fig:teaser}, consists of a first sub-network, referred to as the \emph{student} network, which receives as input an image from a camera view and predicts a disparity map such as to recover the opposite view. On top of this 
network, we propose several contributions. First, from the generated image, we propose to  re-synthesize the input image by estimating the opposite disparity. The resulting network forms a cycle. Second, a third network exploits the cycle
inconsistency between the original and the reconstructed input images in order to refine the estimated depth maps. Our intuition is that inconsistency maps provide rich information which can be further exploited, as they indicate where the first two networks fail to predict disparity pixels. Finally, we propose to use the principle of distillation in order to transfer knowledge from the whole network, seen as a \emph{teacher}, to the \emph{student} network. 
Interestingly, our framework produce two outputs, corresponding to the depth maps estimated respectively by the \textit{student} and the \textit{teacher} networks. This is extremely relevant in practical applications, as the \emph{student} network can be exploited 
in case of low computation power or real-time constraints. 

Our extensive experiments on two large publicly available datasets, \ie~the~KITTI \cite{Geiger2013IJRR} and the Cityscapes \cite{Cordts2016Cityscapes} datasets, demonstrate the effectiveness of the proposed
framework. Notably, by combining the proposed cycle structure with our inconsistency-aware refinement, our unsupervised framework outperforms previous usupervised approaches, while obtaining comparable results with the state-of-the-art supervised methods on the KITTI dataset. 


\begin{figure*}[!t]
\vspace{-10pt}
\begin{center}
\resizebox{.98\textwidth}{!}{
\includegraphics[width=0.90\textwidth]{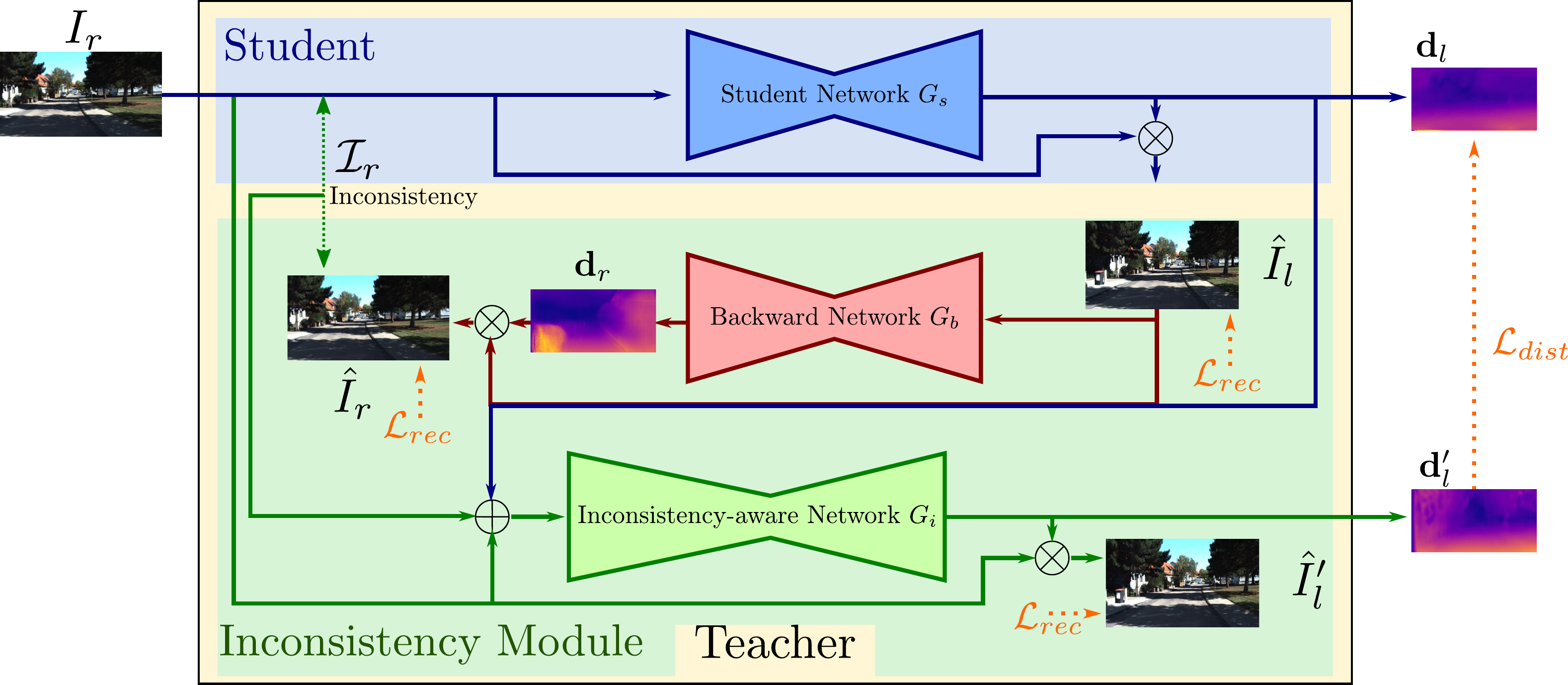}
}
\caption{The proposed approach is composed of two modules. A first network $G_s$ predicts the right-to-left disparity map $d_l$ from the right image and synthesizes the left image as described in Sec.~\ref{sub:optimization}. In the second module, a generator network $G_b$ predicts the left-to-right disparity map $d_r$ in order to re-synthesize the right image. The model obtained in this way forms a cycle. The cycle inconsistency is used by a third network to predict the final disparity map. We use a set of losses (orange dot arrows) detailed in Sec.~\ref{sub:optimization}}.
\label{fig:method}
\end{center}
\vspace{-32pt}
\end{figure*}

\section{Related Work}
\label{sec:related}
In the last decade, deep learning models have greatly
improved the performance of depth estimation methods. The vast majority of methods focus on a supervised setting and the problem of predicting depth maps is cast as a pixel-level regression problem
~\cite{eigen2014depth,liu2016learning,zhuo2015indoor,laina2016deeper,xu2018monocular,Yang2018ECCV, Fu2018CVPR}. The first ConvNet approach for monocular depth prediction was proposed in Eigen~\etal~\cite{eigen2014depth}, where the benefit of considering both local and global information was demonstrated. More recent works improved the performance of deep models by exploiting probabilistic graphical models implemented
as neural networks \cite{liu2016learning,wang2015towards,xu2018structured,xu2018monocular}. 
For instance, Wang~\etal~\cite{wang2015towards} proposed integrating hierarchical Conditional Random Fields (CRFs) into a ConvNet for joint depth estimation and semantic segmentation. Xu~\etal~\cite{xu2018structured,xu2018monocular} exploited CRFs within a deep architecture in order to fuse information at multiple scales. However, supervised approaches rely on expensive ground-truth annotations and, consequently, lack flexibility for deployment in novel environments.


Recently, several works proposed to tackle the depth estimation problem within an unsupervised learning framework~\cite{kuznietsov2017semi,mahjourian2018unsupervised,wang2017learning,zhan2018unsupervised,pilzer2018unsupervised}. 
For instance, Garg~\etal~\cite{garg2016unsupervised} attempted to learn depth maps in an indirect way. They used a ConvNet to predict the right-to-left disparity map from the left image and then reconstructed the right image according to the predicted disparity. They also introduced a network architecture operating based on a
coarse-to-fine principle, \textit{i.e.} they employed an encoder-decoder network where the decoder first estimates a low resolution disparity map and then refines it in order to obtain a map at higher resolution.
Improving upon \cite{garg2016unsupervised}, Godard~\etal~\cite{godard2017unsupervised} 
proposed to use a single generative network to estimate both the left-to-right and the right-to-left disparity maps. Consistency between the two disparities was exploited in form of a loss in order to better constrain the model. 
Other recent works demonstrated that temporal information and, in particular, considering multiple consecutive frames contribute to improve depth estimation ~\cite{wang2018learning,yang2018every,godard2018digging,zhou2017unsupervised}. In particular, Zhou~\etal~\cite{zhou2017unsupervised} exploited temporal information to jointly learn the depth and the camera ego-motion from monocular sequences. Similarly, in \cite{godard2018digging}, a deep network was designed in order to estimate both the depth and the camera pose from three consecutive frames.
In this paper we focus on improving \textit{frame-level} unsupervised depth estimation and we do not exploit any additional information such as supervision from related tasks (\eg ego-motion estimation) or temporal consistency. In this respect, our work can be regarded as complementary to \cite{zhou2017unsupervised,godard2018digging}.

The idea of exploiting cycle-consistency for depth estimation was recently investigated in \cite{pilzer2018unsupervised}. Specifically, Pilzer~\etal~\cite{pilzer2018unsupervised} introduced a deep architecture for stereo depth estimation which is organized in form of a cycle: two sub-networks, corresponding to the two half-cycles, estimate respectively the left-to-right and right-to-left disparities.
They also showed that cycle consistency, together with an adversarial loss, can greatly improve the quality of the predicted depth maps. The main difference with our proposal is that the architecture in \cite{pilzer2018unsupervised} is designed for stereo depth estimation whereas we focus on the monocular setting. Moreover, contrary to \cite{pilzer2018unsupervised}, our architecture exploits cycle inconsistency both at training and at test time. 
Simultaneously, Tosi~\etal~\cite{Tosi_2019_CVPR} proposed disparity refinement and Yang~\etal~\cite{Yang2018ECCV} proposed to compute the error maps between the original input images and their cycle-reconstructed versions and considered them as an additional input to a second network which produces refined depth estimates. 
Opposite to our approach, the deep model in \cite{Yang2018ECCV} is trained using supervision derived by Stereo Direct Sparse Odometry \cite{wang2017stereo}. Furthermore, to construct the cycle, we exploit a backward network and introduce a distillation loss. 

Recently, knowledge distillation attracted a lot of attention \cite{hinton2015distilling}. This methodology consists in compressing a large deep network (usually referred to as the \emph{teacher}) into a much smaller model (\emph{student}) operating on the same modality. The \emph{student} network is trained such that its outputs match those of the \emph{teacher}. Knowledge distillation has been exploited for many computer vision tasks such as domain adaptation \cite{gupta2016cross}, object detection \cite{chen2017learning}, learning from noisy labels~\cite{li2017learning} or facial analysis~\cite{luo2016face}. However, to the best of our knowledge, this work is the first attempt to exploit distillation for depth estimation. We claim that distillation is especially relevant for depth estimation since, in practical applications such as autonomous driving, real-time constraints may impose limitations in term of network size.
Note that, we employ an unusual distillation scenario in which the \emph{student} network is a sub-network of the \emph{teacher}.
\section{Proposed Method
}
\label{sec:method}

\subsection{Overview}
The aim of this work is to estimate the depth of a scene from a single image.
However, at training time, we consider that we dispose of pairs of images $\{\vect{I}_l, \vect{I}_r\}$ of size $H\times W$, derived from a stereo pair and corresponding to the same time instant. Here, $\vect{I}_l$ denotes the left camera view and $\vect{I}_r$ is the right camera view. Given $\vect{I}_r$, we are interested in predicting a correspondence map $\vect{d}_l\in \mathbb{R}^{H\times W}$, namely the right-to-left disparity, in which each pixel value represents the offset of the corresponding pixel between the right and the left images. Finally, assuming that the images are rectified, the depth at a pixel location $(x,y)$ of the left image can be recovered from the predicted disparity with $d_l=\frac{f.b}{d(x,y)}$, where $b$ is the distance between the two cameras and $f$ is the camera focal length.

An overview of the proposed framework is shown in Fig.~\ref{fig:method}. A first network $G_s$ predicts the right-to-left disparity map $\vect{d}_l$ from the right image $\vect{I}_r$, and synthesizes the left image by warping $\vect{I}_r$ according to $\vect{d}_l$. Roughly speaking, the network $G_s$ is trained to minimize the discrepancy between the real and the reconstructed left image (Sec. \ref{sub:optimization}). 

We employ a second generator network $G_b$ that takes as input the synthesized left image and predicts a left-to-right disparity map $\vect{d}_r$ that is used to re-synthesize the right image. The model obtained in this way forms a cycle. This cycle design has three advantages. First, at training time, by sharing weights between $G_s$ and $G_b$, the networks learn to predict disparity maps
from the images of the training set (in the forward half-cycle $G_s$) but also from the
synthesized images (in the backward half-cycle $G_b$). In that sense, the use of the cycle can be seen as a sort of data augmentation. 
Second, in order to re-synthesize correctly the right image, the second network $G_b$ requires a correct input left image. Thus, $G_b$ imposes a global constraint on the estimated disparity 
$\vect{d}_l$ 
oppositely to standard pixel-wise discrepancy losses, such as $\mathcal{L}_1$ or $\mathcal{L}_2$ that act only locally. 
Third, by comparing the input right image $\vect{I}_r$ and the output right image $\hat{\vect{I}_r}$ synthesized after applying our cycle framework, we can measure the cycle inconsistency. At a given location of the input image, if we observe no inconsistency, $G_s$ and $G_b$ must have predicted correctly the disparity maps. Conversely, in case of inconsistency, $G_s$ or $G_b$ (or both) must have predicted incorrectly the disparity maps. 
Note that inconsistencies may also appear on objects regions that are visible in only one of the two views. Interestingly, these regions are usually located on the object edges. Therefore, looking at cycle inconsistency also provides information about object edges that can help to predict better depth maps. 
Importantly, this inconsistency can be measured both at training and testing times, even if at testing time, we dispose only of the right image. 

The main contribution of this work consists in exploiting the cycle inconsistency by training a third network in order to improve the prediction performance and output a refined depth map $\vect{d}_l'$. In addition, since employing our inconsistency-aware network leads to more accurate depth predictions, we propose to use the disparity maps predicted by $G_i$ in order to improve $G_s$ training via a knowledge distillation approach.

Note that, another possible cycle approach, as proposed in \cite{Yang2018ECCV}, would consist in using a single network to predict the two disparity maps. The two disparities can be used to obtain the synthesized left image and then the re-synthesized right image.
Nevertheless, this approach has a major disadvantage with respect to our approach,~\ie, 
since only the warping operator in employed between the two synthesized images, and consequently the receptive field of $\hat{\vect{I}_r}$ in $\vect{I}_l$ is very small. In particular, when implementing the warping operator via bilinear sampling, the receptive field of the warping operator in only $2\times2$. Therefore, the right image reconstruction loss can act on the reconstructed left image only locally. Conversely, our backward network $G_b$ imposes a global consistency on $\vect{d}_l$ thanks to its large receptive field.

The outputs of our method correspond to the estimated depth maps $\vect{d}_l$ and $\vect{d}_l'$.
While the estimated depth $\vect{d}_l'$ corresponding to the teacher model is typically more accurate, in some applications, \eg in resource-constrained settings, it could be convenient to exploit only a small student network.

In the following, we describe the design of our cycled network. Then, we introduce our novel inconsistency-aware network. Finally, we present the optimization objective including our proposed distillation approach.

\subsection{Unsupervised Monocular Cycled Network}
\label{sub:cyclenet}

In this work, we adopt a setting in which the model is trained without the need of ground truth depth maps. 
This approach is often referred to as unsupervised or self-supervised depth estimation. Roughly speaking, it consists in training a network to predict a disparity map that can be used to generate the left image from the right image.  
Formally speaking, we employ a first network $G_s$ that takes as input the right image $\vect{I}_r$ and predicts the right-to-left disparity $\vect{d}_l$. Following~\cite{godard2017unsupervised}, we adopt a U-Net architecture for $G_s$.
We employ a warping function $f_w(\cdot)$ that synthesizes the left view image by sampling from $\vect{I}_r$ according to $\vect{d}_l$:
\begin{equation}
  \hat{\vect{I}}_l = f_w(\vect{d}_l, \vect{I}_r).\label{eq:samp}
\end{equation}
Importantly, $f_w(\cdot)$ is implemented using the bilinear sampler from the spatial transformer network \cite{jaderbergNips15} resulting in a fully differentiable model. Consequently, the network can be trained via gradient descent by minimizing the discrepancy between $\hat{\vect{I}}_l$ and $\vect{I}_l$ (see Sec. \ref{sub:optimization} for details about network training). 

Inspired by \cite{pilzer2018unsupervised}, we employ a second network $G_b$ in order to re-synthesize the right image according to:
\begin{equation}
    \hat{\vect{I}}_r = f_w(\vect{d}_r, \vect{\hat{I}}_l). \label{eq:warp2right}
\end{equation}
where:
\begin{equation}
  \vect{d}_r=G_b(\hat{\vect{I}}_l)
\end{equation}
The $G_b$ and $G_s$ networks share their encoder parameters. Note that, differently from the stereo depth model proposed in \cite{pilzer2018unsupervised}, our second half-cycle network takes only the synthesized left image as input. This crucial difference allows the use of this cycle in the monocular setting at testing time.
Concerning the decoder networks, we adopt an architecture composed of a sequence of up-convolution layers in which the disparity is estimated and gradually refined from low to full resolutions similarly to \cite{godard2017unsupervised}. We obtain the estimated left and the right disparity maps at each scale $\vect{d}_l^n$ and $\vect{d}_r^n$, $n\in\{0,1,2,3\}$, with sizes $[H/2^n,W/2^n]$. 
More precisely, $\vect{d}_r^n$ is computed from the decoder feature map $\vect{\xi_r}^n$ of size $[H/2^n,W/2^n]$ via a convolutional layer. Then, $\vect{d}_r^n$ is concatenated with $\vect{\xi_r}^n$ obtaining a tensor that is input to an up-convolution layer in order to estimate the disparity at the next resolution $\vect{d}_r^{n-1}$.

\subsection{Inconsistency-Aware Network}
\label{sec:IAnet}
We define the inconsistency tensor as the difference between the input image $\vect{I}_r$ and the image $\hat{\vect{I}}_r$ predicted by the backward network $G_b$:
\begin{equation}
  \mathcal{I}_r=\vect{I}_r-\hat{\vect{I}}_r
\end{equation}
The proposed inconsistency-aware network $G_i$ takes as input the concatenation of $\vect{I}_r$, $\mathcal{I}_r$ and $\vect{d}_l$.  
We employ a network architecture similar to the half-cycle monocular network described in Sec. \ref{sub:cyclenet}.
However, we propose to provide to the encoder network the disparity maps $\vect{d}_l^n, n\in\{1,2,3\}$ estimated by $G_s$ at each scale. 
More precisely, we concatenate along the channel axis each disparity $\vect{d}_l^n$ with network features of corresponding dimensionality. 

The inconsistency-aware network $G_i$ estimates the right-to-left disparity $\vect{d}_l' =G_i(\vect{I}_r, \mathcal{I}_r, \vect{d}_l,\vect{d}_l^{\{1,2,3\}})$ and we reconstruct the left view image $\hat{\vect{I}_l}'$ by applying the warping function $f_w$:
\begin{equation}
    \hat{\vect{I}}_l' = f_w(\vect{d}_l', \vect{I}_r)
\end{equation}
Similarly to $G_s$ and $G_b$, $G_i$ estimates low resolution disparity maps $\vect{d_l'}^{n}, n\in\{1,2,3\}$ that are gradually refined from low to full resolutions. 

\subsection{Network Training and Knowledge Self-Distillation}
\label{sub:optimization}
In this section, we detail the losses employed to train the proposed network in an end-to-end fashion.

\noindent
\textbf{Reconstruction.} First, we employ a reconstruction and stucture similarity loss for each network. Following \cite{godard2017unsupervised}, we adopt the $\mathcal{L}_1$ loss to measure the discrepancy between the synthesized and the real images and the structure similarity loss $\mathcal{L}_{SSIM}$ to measure the discrepancy between the synthesized and the real images structure. By summing the losses of the three networks $G_s$, $G_b$ and $G_i$, we obtain:
\begin{equation}
\begin{split}
    \mathcal{L}_{rec}^{(0)} = \lambda_s[\alpha\mathcal{L}_{SSIM}(\vect{\hat{I}}_l,\vect{I}_l) + (1-\alpha)||\vect{\hat{I}}_l - \vect{I}_l||_1] \\ + \lambda_b [\alpha\mathcal{L}_{SSIM}(\vect{\hat{I}}_r,\vect{I}_r) + (1-\alpha)||\vect{\hat{I}}_r - \vect{I}_r||_1] \\ + \lambda_t[ \alpha\mathcal{L}_{SSIM}(\vect{\hat{I}}_l',\vect{I}_l) + (1-\alpha) ||\vect{\hat{I}}_l' - \vect{I}_l||_1]
\end{split}
\end{equation}
where $\lambda_s$, $\lambda_b$ and $\lambda_t$ are adjustment parameters and $\alpha = 0.85$. 
Similarly, we also compute a reconstruction loss $\mathcal{L}_{rec}^{(n)}$  for the low resolution disparity maps. Following \cite{godard2018digging}, we upsample the low resolution $\vect{d}_l^n$, $\vect{d}_r^n$ and $\vect{d_l'}^{n}$ to $H\times W$ and use the warping operator $f_w$ to re-synthesize full resolution images that are compared with the real images according to the $\mathcal{L}_1$ loss.
The total reconstruction loss is:
\begin{equation}
    \mathcal{L}_{rec} = \sum_{n=0}^4 \mathcal{L}_{rec}^{(n)}
\end{equation}

 
\noindent
\textbf{Self-Distillation.}
Finally, we propose to introduce a knowledge distillation loss. As detailed in the experimental section (Sec \ref{sec:exp}), the inconsistency-aware network outperforms by a significant margin the simple half-cycle network $G_s$. This boost is at the cost of a higher computation complexity. The idea of the proposed self-distillation loss consists in distilling knowledge from inconsistency-aware network to the half-cycle network $G_s$. Thus, we improve the performance of $G_s$ without adding any computation complexity at testing time. To do so, we evaluate disparity and feature distillation. For the first, we impose that the network $G_d$ predicts disparity maps similar to the output of inconsistency-aware network. It can be seen as a distillation approach where $G_s$ plays the role of the \emph{student} and the whole network (composed of $G_s$, $G_b$ and $G_i$) is the \emph{teacher}. However, in our particular case, the \emph{student} network is a sub-network of the \emph{teacher}. From this perspective, we name this approach self-distillation. The self-distillation loss is given by:

\begin{equation} \label{eq:loss-disp-dist}
    \mathcal{L}_{dist} = ||\vect{d}_l - \mathcal{S}(\vect{d}_l')||_1
\end{equation}
where $\mathcal{S}$ denotes the stop-gradient operation. In particular, the stop-gradient operation equals the identity function when computing the forward pass of the back-propagation algorithm but it has a null gradient when computing the backward pass. The purpose of the stop-gradient is to avoid that $\vect{d}_l'$ converges to $\vect{d}_l$. On the contrary, the goal is to help $\vect{d}_l$ to become as accurate as $\vect{d}_l'$.

For the second, we impose that the decoder features $\xi_r'^n, n\in {0,1,2}$ of the \textit{teacher} are similar to the features $\xi_r^n$ of the \textit{student}. The self-distillation loss is given by:
\begin{equation}
    \mathcal{L}_{dist} = || \xi_r^n - \mathcal{S}(\xi_r'^n) ||_2
\end{equation}

The total training loss is given by: 
\begin{equation} \label{eq:loss-total}
    \mathcal{L}_{tot} = \mathcal{L}_{rec}+\lambda_{dist}\mathcal{L}_{dist}
\end{equation}

\section{Experiments}
\label{sec:exp}

We evaluate our proposed approach on two publicly available datasets and compare its performance with state of the art methods.

\subsection{Experimental Setup}
\textbf{Datasets.} We perform experiments on two large stereo images datasets, \ie~KITTI~\cite{kitti} and Cityscapes~\cite{Cityscapes}. Both datasets are recorded from driving vehicles. Concerning the \emph{KITTI} dataset, we employ the training and test split of Eigen \etal~\cite{eigen2014depth}. This split is composed of 22,600 training image pairs, and 697 test pairs. We consider data-augmentation with online random flipping of the images during training as in \cite{godard2017unsupervised}. 
For Cityscapes
, images were collected with higher resolution. To train our model we combine images from the densely and coarse annotated splits to obtain 22,973 image-pairs as in \cite{pilzer2018unsupervised}. The test split is composed of 1,525 image-pairs of the densely annotated split. The evaluation is performed using the pre-computed disparity maps.

\noindent
\textbf{Evaluation Metrics.} The quantitative evaluation is performed according to several standard metrics used in previous works~\cite{eigen2014depth, godard2017unsupervised, wang2015towards}. Let $P$ be the total number of pixels in the test set and $\hat{d}_i$, $d_i$ the estimated depth and ground truth depth values for pixel $i$. We compute the following metrics: 
\begin{itemize}[noitemsep]\setlength
\item 
 Mean relative error (abs rel): 
$\frac{1}{P} \sum_{i=1}^{P} \frac{\parallel \hat{d}_i - d_i \parallel}{d_i}$, 
\item Squared relative error (sq rel): 
$\frac{1}{P} \sum_{i=1}^{P} \frac{\parallel \hat{d}_i - d_i \parallel^2}{d_i}$, 
\item Root mean squared error (rmse): 
$\sqrt{\frac{1}{P}\sum_{i=1}^P(\hat{d}_i - d_i)^2}$, 
\item Mean $\log10$ error (rmse log):
$\sqrt{\frac{1}{P} \sum_{i=1}^{P} \parallel \log \hat{d}_i - \log d_i \parallel^2}$
\item Accuracy with threshold $\tau$, \ie the percentage of $\hat{d}_i$ such that $\delta = \max (\frac{d_i}{\hat{d}_i},\frac{\hat{d}_i}{d_i}) < \alpha^\tau$. We employ $\alpha = 1.25$ and $\tau \in [1,2,3]$ following \cite{eigen2014depth}.
\end{itemize}



\subsection{Baselines for Ablation.}
To perform the ablation study presented in Sec.\ref{sub:abl}, we consider the following baselines:
\begin{itemize}[noitemsep]\setlength
\itemsep{-0.1em}
\item \textit{half-cycle}: our basic building block, uses the forward branch that takes $\vect{I}_r$ as input and generates $\vect{d}_l$ to reconstruct the other stereo view $\vect{\hat{I}}_l$. Neither cycle-consistency nor self-distillation are used in this model.
\item \textit{cycle}: a backward network is added to the \textit{half-cycle} model in order to reconstruct $\vect{\hat{I}}_r$ from the estimated $\vect{\hat{I}}_l$. Note that the backward network is used only at training time. At test time, the output is the same as for the \textit{half-cycle} model.
\item \textit{teacher}, we stack the inconsistency-aware network after the \textit{cycle} as described in Sec \ref{sec:IAnet}.
\item \textit{student}: the output of the inconsistency-aware network is distilled in order to refine the first \textit{half-cycle}. At test time, the output and the computation complexity are the same as in the \textit{half-cycle} model.
\end{itemize}
In Tables~\ref{tab:new_kitti_shit},~\ref{tab:dsvo} and ~\ref{tab:ablation_city} we indicate with \textit{HC}, \textit{C}, \textit{T} and \textit{S}, the \textit{half-cycle}, \textit{cycle}, \textit{teacher} and \textit{student} respectively; \textit{feat} and \textit{disp} denote self-distillations of features and disparities.

\noindent
\textbf{Training Procedure.} The whole network is trained following an iterative procedure. First, we start by training the forward \textit{half-cycle} network for 10 epochs. In a second step, we train the backward network decoder for 5 epochs without updating the first half-cycle network. The whole cycle is then jointly trained for further 10 epochs. Then, the inconsistency-aware module is pretrained for 5 epochs. Finally, the whole network is jointly fine-tuned for 10 epochs.

\noindent
\textbf{Parameters.} 
The model is implemented with the deep learning library \textit{TensorFlow}. Similarly to \cite{godard2017unsupervised}, the input images are down-sampled to a resolution of $512\times 256$ from the original sizes which are $1226\times 370$ for the KITTI dataset and for CityScapes. 
In all our experiments we use a batch size equal to $8$ stereo image pairs and the Adam optimizer with learning rate set to $10^{-5}$.

The \textit{half-cycle} and \textit{cycle} networks are trained  with the following loss parameters $\lambda_s = 1$, $\lambda_b = 0.1$ and $\lambda_t = 0$. 
When training the \textit{teacher} network we use $\lambda_s = 0$, $\lambda_b = 0$ and $\lambda_t = 1$. We weight the distillation loss $\mathcal{L}_{dist}$ with $\lambda_{dist} = 0.005$ and $\lambda_{dist} = 0.1$ respectively, if feature distillation or disparity distillation is applied.
The joint training of the full network is done with learning rate $l_r = 10^{-5}$, loss parameters $\lambda_s = 1$, $\lambda_b = 0.1$, $\lambda_t = 1$ and $\lambda_{dist}$ equal to $0.005$ in the case feature distillation and $0.1$ in the case of disparity distillation, respectively.

\noindent

\begin{table}[t!]
\centering
\resizebox{.999\columnwidth}{!}{
\begin{tabular}{ | l || c | c | c | c | c | c | c | c |}
\toprule
   \multirow{2}{*}{Method} & Abs Rel & Sq Rel & RMSE & $RMSE_{log}$ & $\delta<1.25$ & $\delta<1.25^2$ & $\delta<1.25^3$ \\
   \cline{2-5} \cline{6-8}
    & \multicolumn{4}{c|}{lower is better}& \multicolumn{3}{c|}{higher is better} \\
\midrule
HC &0.1487 &1.2942 &5.800 &0.246 &0.805 &0.925 &0.965 \\
C &0.1451 &1.2943 &5.850 &0.242 &0.796 &0.924 &0.967\\
\midrule
T \textit{feat} &\textit{0.1220} &\textit{1.0433} &5.321 &0.229 &\textit{0.834} &0.933 &\textit{0.968}\\
T \textit{disp} &0.1234 &1.0509 &\textit{5.283} &\textit{0.228} &\textit{0.834} &\textit{0.934} &\textit{0.968} \\
\midrule
S \textit{feat} &0.1438 &1.2806 &5.834 &0.241 &0.797 &0.926 &0.968 \\
S \textit{disp} &0.1438 &1.2551 &5.771 &0.238 &0.797 &0.927 &0.969 \\
\midrule
\multicolumn{8}{|c|}{\cite{godard2018digging} $\mathcal{L}_1$ loss}\\
\midrule
T \textit{feat} &0.1017 &0.8930 &4.768 &0.206 &0.878 &0.946 &0.972\\ 
T \textit{disp} &\textbf{0.0983} &\textbf{0.8306} &\textbf{4.656} &\textbf{0.202} &\textbf{0.882} &\textbf{0.948} &\textbf{0.973} \\ 
\midrule
S \textit{feat} &0.1474 &1.2416 &5.849 &0.241 &0.788 &0.923 &0.968 \\ 
S \textit{disp} &0.1424 &1.2306 &5.785 &0.239 &0.795 &0.924 &0.968 \\ 
\bottomrule
\end{tabular}
}

\caption{Ablation study on KITTI dataset using the training and testing split proposed by Eigen~\etal~\cite{eigen2014depth}. The upper part shows the results with the multiscale reconstruction $\mathcal{L}_1$ loss in~\cite{godard2017unsupervised}, the bottom part with the $\mathcal{L}_1$ loss proposed in~\cite{godard2018digging}. 
}
\label{tab:new_kitti_shit}
\end{table}

\begin{table}[t]
\vspace{-8pt}
\centering
\resizebox{.999\columnwidth}{!}{
\begin{tabular}{ | l || c | c | c | c | c | c | c |}
\toprule
   \multirow{2}{*}{Method} & Abs Rel & Sq Rel & RMSE & $RMSE_{log}$ & $\delta<1.25$ & $\delta<1.25^2$ & $\delta<1.25^3$ \\
   \cline{2-5} \cline{6-8}
    & \multicolumn{4}{c|}{lower is better}& \multicolumn{3}{c|}{higher is better} \\

\midrule
1-CN C &0.1533 &1.3326 &5.837 &0.240 &0.785 &0.919 &0.967 \\

1-CN S \textit{disp} &0.1503 &1.2622 &5.868 &0.243 &0.783 &0.918 &0.967 \\
Ours S \textit{disp} &\bf0.1438 &\bf1.2551 &\bf5.771 &\bf0.238 &\bf0.797 &\bf0.927 &\bf0.969\\
\midrule
1-CN T \textit{disp} &0.1478 &1.3609 &5.952 &0.243 &0.793 &0.921 &0.966 \\
Ours T \textit{disp} &\bf0.1234 &\bf1.0509 &\bf5.283 &\bf0.228 &\bf0.834 &\bf0.934 &\bf0.968 \\
\bottomrule
\end{tabular}
}
\caption{Ablation study where our two-network \textit{cycle} is replaced by the single-network \textit{cycle} from Yang~\etal~\cite{Yang2018ECCV} (referred as to 1-CN). 
}
\label{tab:dsvo}
\end{table}

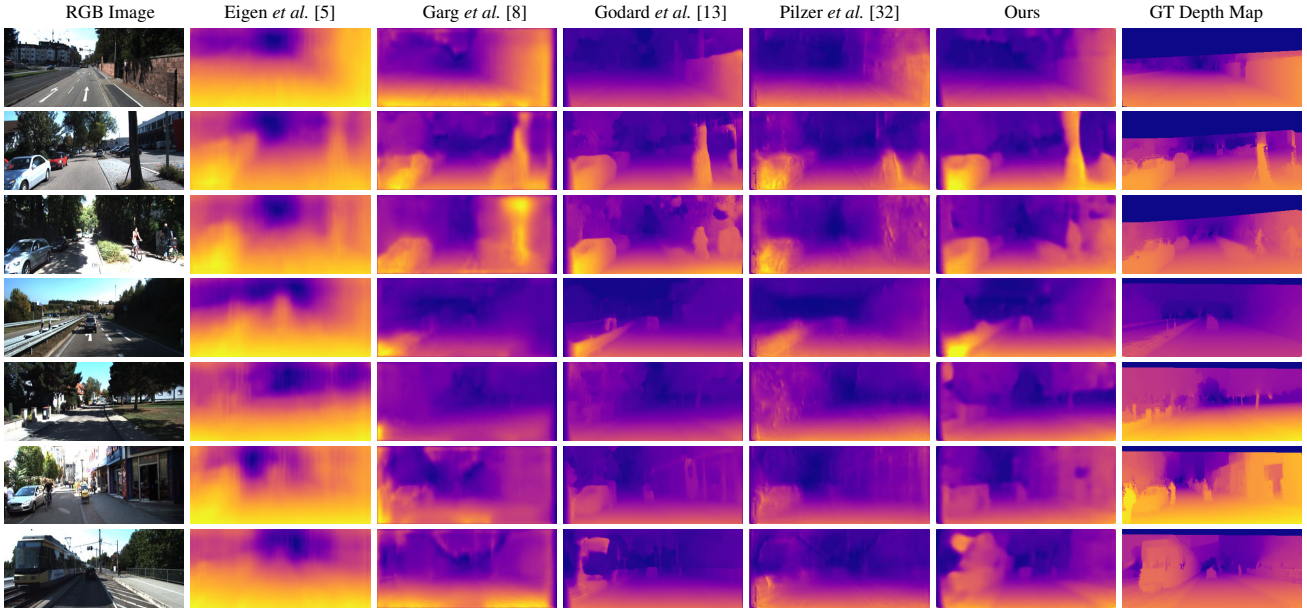
\begin{figure*}
\vspace{-10pt}
	\centering
	\includegraphics[width=\textwidth]{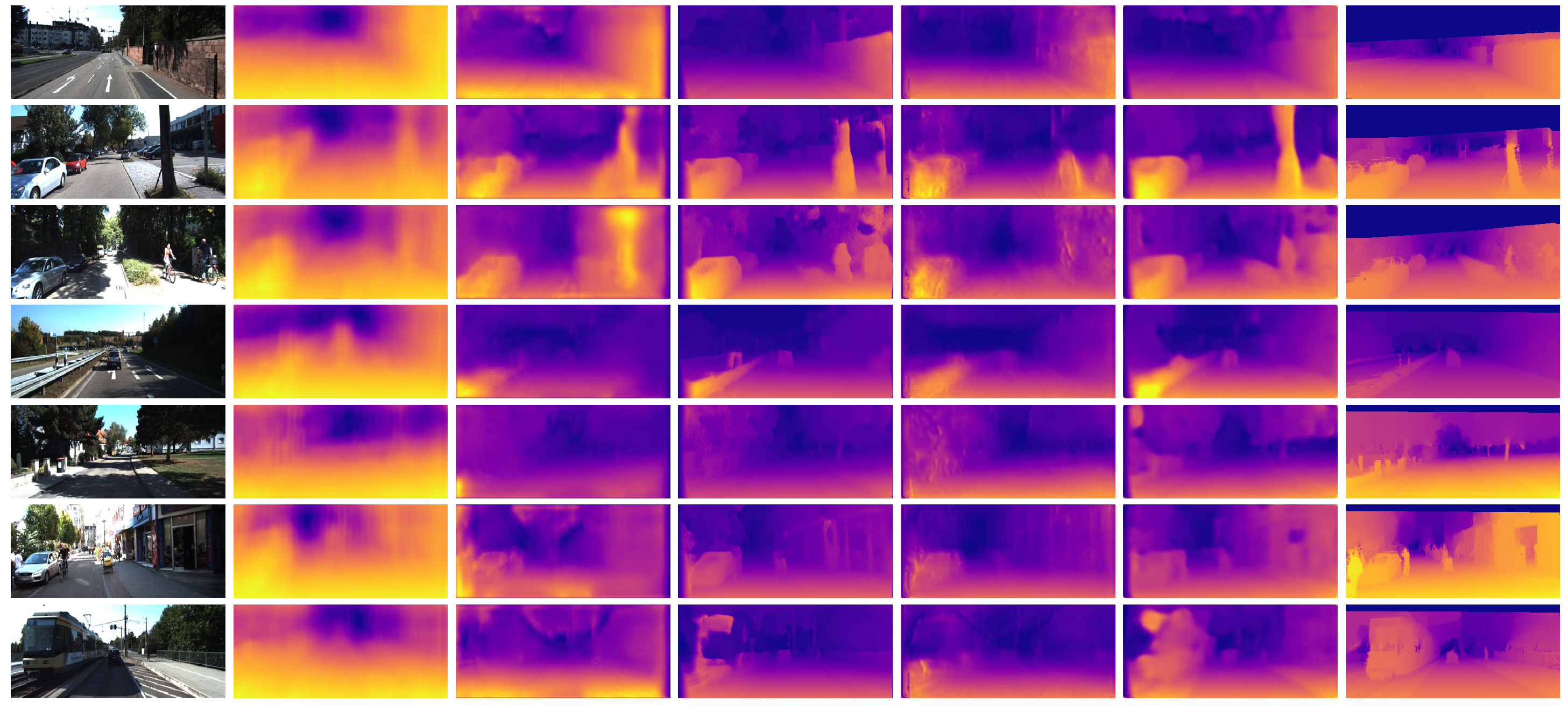}
	\put(-470,227){\scriptsize RGB Image}
	\put(-410,227){\scriptsize Eigen~\etal~\cite{eigen2014depth}}
	\put(-335,227){\scriptsize Garg~\etal~\cite{garg2016unsupervised}}
	\put(-270,227){\scriptsize Godard~\etal~\cite{godard2017unsupervised}}
	\put(-200,227){\scriptsize Pilzer~\etal~\cite{pilzer2018unsupervised}}
	\put(-115,227){\scriptsize Ours}
	\put(-60,227){\scriptsize GT Depth Map}
	\caption{Qualitative comparison of different state-of-the-art models with our \textit{teacher} network on the KITTI testing split proposed by~\cite{eigen2014depth}. The sparse KITTI ground truth depth maps are interpolated with bilinear interpolation for better visualization.}
	\label{fig:kitti_sota}
\vspace{-4pt}
\end{figure*}

\begin{table}[ht]
\vspace{-8pt}
\centering
\resizebox{.999\columnwidth}{!}{
\begin{tabular}{ | l || c | c | c | c | c | c | c |}
\toprule
   \multirow{2}{*}{Method} & Abs Rel & Sq Rel & RMSE & $RMSE_{log}$ & $\delta<1.25$ & $\delta<1.25^2$ & $\delta<1.25^3$ \\
   \cline{2-5} \cline{6-8}
    & \multicolumn{4}{c|}{lower is better}& \multicolumn{3}{c|}{higher is better} \\
\midrule
HC &0.4676 & 7.3992 & 5.741 & 0.493 & 0.735 & 0.890 & 0.945 \\
C &0.4523 &6.2604 &5.381 &0.557 &0.736 &0.888 &0.946\\ 
\midrule
T \textit{feat}&0.4087 &5.8777 &4.394 &0.334 &0.846 &0.940 &0.967\\ 
T \textit{disp}&\textit{0.3988} &\textit{5.8752} &\textit{4.293} &\textit{0.316} &\textit{0.848} &\textit{0.941} &\textit{0.968}\\ 
\midrule
S \textit{feat}&0.4494 &6.2599 &5.343 &0.421 &0.739 &0.891 &0.947\\
S \textit{disp}&0.4467 &5.9012 &5.297 &0.473 &0.736 &0.890 &0.946\\
\midrule
\multicolumn{8}{|c|}{\cite{godard2018digging} $\mathcal{L}_1$ loss}\\
\midrule
T \textit{feat}&0.3878 &\textbf{5.8190} &\textbf{4.123} &0.397 &0.861 &\textbf{0.945} &\textbf{0.969}\\
T \textit{disp}&\textbf{0.3846} &6.2007 &4.476 &\textbf{0.318} &\textbf{0.864} &\textbf{0.945} &\textbf{0.969}\\
\midrule
S \textit{feat}&0.4455 &6.2748 &5.366 &0.468 &0.739 &0.891 &0.946\\
S \textit{disp}&0.4305 &5.9552 &5.281 &0.519 &0.740 &0.891 &0.946\\
\bottomrule
\end{tabular}
}
\caption{Ablation study on the Cityscapes dataset. 
 The upper part shows the results with the multiscale reconstruction $\mathcal{L}_1$ loss in~\cite{godard2017unsupervised}, the bottom part with the $\mathcal{L}_1$ loss proposed in~\cite{godard2018digging}.
}
\label{tab:ablation_city}
\vspace{-12pt}
\end{table}

\begin{table*}
	\begin{center}
	\resizebox{.9\textwidth}{!}{
		\begin{tabular}{ | l | c | c || c | c | c | c | c | c | c |}
			\toprule
			\multirow{2}{*}{Method} & \multirow{2}{*}{Sup}&\multirow{2}{*}{Video} & Abs Rel & Sq Rel & RMSE & $RMSE_{log}$ & $\delta<1.25$ & $\delta<1.25^2$ & $\delta<1.25^3$ \\
			\cline{4-7} \cline{8-10}
			&&& \multicolumn{4}{c|}{lower is better}& \multicolumn{3}{c|}{higher is better} \\
			\midrule
			Eigen~\etal~\cite{eigen2014depth}&Y &N &0.190&1.515&7.156&0.270&0.692&0.899&0.967\\
			Xu~\etal~\cite{xu2018monocular}  & Y  &N & 0.132 & 0.911 & - & \textit{0.162} &0.804 &0.945 & 0.981 \\
            Jiang~\etal \cite{Jiang2018eccv} &Y &N  &0.131 &0.937 &5.032 &0.203 &0.827 &0.946 &0.981 \\
            Gan~\etal~\cite{Gan2018ECCV} &Y &N  &0.098 &0.666 &\textit{3.933} &0.173 &\textit{0.890} &0.964 &0.985 \\
            Guo~\etal~\cite{Guo2018ECCV} &Y &N &\textit{0.097} &\textit{0.653} &4.170 &0.170 &0.889 &\textit{0.967} &\textit{0.986} \\
						\midrule
						\midrule
			Yang~\etal~\cite{Yang2018ECCV} &Y &Y &\textit{0.097} &\textit{0.734} &\textit{4.442} &\textit{0.187} &\textit{0.888} &\textit{0.958} &\textit{0.980} \\
			Zou~\etal\cite{DFNet2018eccv} &N& Y &0.150 &1.124 &5.507 &0.223 &0.806 &0.933 &0.973 \\
			Godard~\etal \cite{godard2018digging} &N & Y &0.115 &1.010 &5.164 &0.212 &0.858 &0.946 &0.97 \\
			\midrule
									\midrule    
			Zhou~\etal~\cite{zhou2017unsupervised}&N &N &0.208&1.768&6.856&0.283&0.678&0.885&0.957\\
			Garg~\etal~\cite{garg2016unsupervised}&N &N &0.169&1.08&5.104&0.273&0.740&0.904&0.962\\
			Kundu~\etal~\cite{adadepth}, 50m&N &N &0.203&1.734&6.251&0.284&0.687&0.899&0.958\\
			Godard~\etal~\cite{godard2017unsupervised}&N &N &0.148&1.344& 5.927& 0.247 &0.803&0.922 & 0.964\\
			Pilzer~\etal~\cite{pilzer2018unsupervised} & N  &N &0.152 &1.388 &6.016 &0.247 &0.789 &0.918 &0.965\\
			\midrule    
			Ours Student & N  &N &0.1424 &1.2306 &5.785 &0.239 &0.795 &0.924 &0.968\\
			Ours Teacher &N  &N &\textbf{0.0983} &\textbf{0.8306} &\textbf{4.656} &\textbf{0.202} &\textbf{0.882} &\textbf{0.948} &\textbf{0.973}\\
			\bottomrule
		\end{tabular}
	}
	\end{center}
	\caption{Comparison with the state of the art. Training and testing are performed on the KITTI \cite{kitti} dataset. Supervised and semi-supervised methods are marked with Y in the supervision (Sup.) column, unsupervised methods with N. Methods using a frame sequence in input and, thus, exploiting temporal information either at training or testing time, are marked with Y in the \emph{Video} column. Numbers are obtained on Eigen~\cite{eigen2014depth} test split with Garg~\cite{garg2016unsupervised} image cropping. Depth predictions are capped at the common threshold of 80 meters, if capped at 50 meters we specify it. Best scores among static unsupervised methods are in bold. Best scores among other method categories are in italic.}
\label{tab:state-art_kitti}
\vspace{-12pt}
\end{table*}

\begin{figure*}
\vspace{-10pt}
	\centering
	\includegraphics[width=\textwidth]{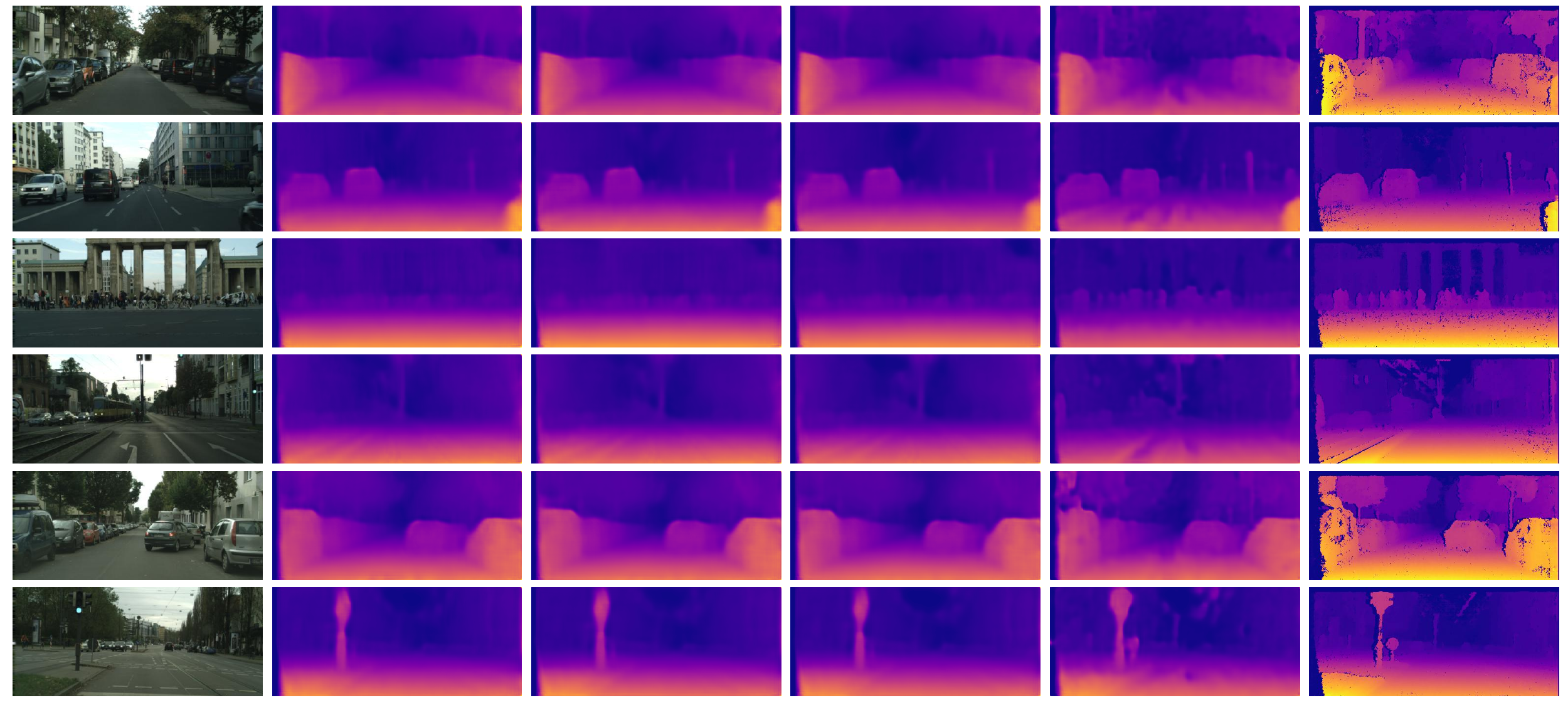}
	\put(-470,227){\scriptsize RGB Image}
	\put(-385,227){\scriptsize \textit{half-cycle}}
	\put(-295,227){\scriptsize \textit{cycle}}
	\put(-220,227){\scriptsize \textit{student}}
	\put(-140,227){\scriptsize \textit{teacher}}
	\put(-65,227){\scriptsize GT Depth Map}
	\caption{Qualitative comparison of different baseline models of the proposed approach on the Cityscapes testing dataset.}
	\label{fig:cityscapes}
\vspace{-12pt}
\end{figure*}

\subsection{Results}
\label{sub:abl}
\noindent
\textbf{Ablation Study.} To demonstrate the validity of the proposed contributions we first conduct an ablation study on the KITTI dataset \cite{kitti} and the CityScapes dataset \cite{Cityscapes}. Results are shown in Table \ref{tab:new_kitti_shit} and Table \ref{tab:ablation_city}, respectively. 

We split the ablation in two parts where we employ two different reconstruction loss variants. For the first part, as in~\cite{godard2017unsupervised}, we use a multi-scale reconstruction loss where the smaller scale reconstruction is compared with a downsampled version of the stereo image. In contrast with that, for the second part, we employ a more effective reconstruction loss, upsampling to input scale all the disparities before warping as described in Sec.~\ref{sub:optimization}.

In Table~\ref{tab:new_kitti_shit} it is interesting to note that our intuition of 
self-constraining the monocular student network with cycled design improves, without requiring additional losses, in several of the metrics compared to the simple forward branch. This comes at the cost of doubling the forward propagation time at training but not at testing time. Moreover, the monocular cycled structure has the big advantage of automatically computing the inconsistency of the reconstruction both at training and testing time. Therefore, stacking a network aware of the inconsistencies and previous estimations, the \textit{teacher} network, improves the performance. We observe that our proposed inconsistency-aware network brings an important improvement consistent over all the metrics, \eg $14\%$ and $18\%$ in \textit{Abs Rel} and \textit{Sq Rel}, respectively, comparing \textit{cycle} and \textit{teacher}.

Student-teacher distillation leads to a consistent improvement over all metrics, demonstrating that self-distillation improves the \textit{student}, while keeping the performance of teacher constant. Regarding the two distillation strategies, we found that network with disparity distillation converges faster than that with the feature distillation. This is not unexpected, given the much more compact size of the disparity compared to the several channels of the features. 

For demonstrating the validity of the design of our cycle network, we perform an ablation study where our two-network \textit{cycle} structure is replaced by the single-network \textit{cycle} proposed by Yang~\etal~\cite{Yang2018ECCV}. In this experiment, we use our proposed inconsistency-aware module to exploit the inconsistency estimated by the single network cycle in~\cite{Yang2018ECCV}. Contrary to \cite{Yang2018ECCV}, we trained the models without supervision in order to compare the two different approaches in the unsupervised setting. We use the $\mathcal{L}_1$ loss from~\cite{godard2017unsupervised} for fair comparison.
Results are reported in Table \ref{tab:dsvo}. We observe that the inconsistency estimates obtained with the single-network cycle of \cite{Yang2018ECCV} are associated with worse performance
with respect to those of our method. 


We also performed an ablation study on the Cityscapes dataset in Table~\ref{tab:ablation_city}, following the evaluation procedure proposed in~\cite{pilzer2018unsupervised}. The results confirm the trends observed on KITTI. The \textit{cycle} network improves over the \textit{half-cycle} in five metrics out of seven. The \textit{teacher}, effectively exploiting inconsistencies, is associated with an improvement on all error metrics (ranging from $7\%$ to 20$\%$). Distillation further provides a boost in performance of about $1.5\%$ to $5\%$. In the second part of the ablation study, the \textit{teacher} further improves its estimations gaining over $20\%$ over the initial \textit{cycle} setting. More interesting is the gain in performance of the \textit{student} that improves from $2\%$ to $5\%$.

In Fig.~\ref{fig:cityscapes}, we present qualitative results for Cityscapes. \textit{half-cycle} and \textit{cycle} images are smooth and do not present artifacts. The \textit{teacher} provides more accurate depth maps with sharper edges for small objects and better background estimations (\eg third row, people in the back). After distillation also the \textit{student} inherits this ability and we observe more detailed predictions compared to the original \textit{cycle}.

\subsection{Comparison with State-of-the-Art}
\label{sub:sota}
In Table~\ref{tab:state-art_kitti} we compare with several state-of-the-art works,
considering both supervised learning-based ( Eigen~\etal~\cite{eigen2014depth}, Xu~\etal~\cite{xu2018monocular}, Jiang~\etal~\cite{Jiang2018eccv}, Gan~\etal~\cite{Gan2018ECCV}, Guo~\etal~\cite{Guo2018ECCV}, Yang~\etal~\cite{Yang2018ECCV}) and unsupervised learning-based (Zhou~\etal~\cite{zhou2017unsupervised}, Garg~\etal~\cite{garg2016unsupervised}, Kundu~\etal~\cite{adadepth}, Godard~\etal~\cite{godard2017unsupervised}, Pilzer~\etal~\cite{pilzer2018unsupervised}, Godard~\etal~\cite{godard2018digging} and Zou~\etal\cite{DFNet2018eccv}) methods.

The \textit{teacher} network reaches state-of-the-art performance for the frame-level unsupervised setting, even improving over the state-of-the-art method that use depth supervision as \cite{xu2018monocular}, and is competitive with those using depth and video clues~\cite{Gan2018ECCV,Guo2018ECCV,Yang2018ECCV}. Note that Yang~\etal~\cite{Yang2018ECCV} consider a similar setting to ours proposing to use errors to refine the depth estimation with a stacked network. Our method has several advantages though: it is unsupervised, it does not consider multiple video frames and it avoids the use of several losses whose hyper-parameters are hard to tune.
Furthermore, as demonstrated by our experiments in Table \ref{tab:dsvo}, our approach adopts a more effective network structure for computing cycle inconsistencies.
The \textit{student} network, after distillation, improves on unsupervised approaches with similar network capacity like~\cite{garg2016unsupervised,godard2017unsupervised,pilzer2018unsupervised} and it is only outperformed by previous unsupervised methods that exploit additional information during training like~\cite{godard2018digging}.

Qualitative results in Figure \ref{fig:kitti_sota} show that our model predicts more accurately challenging areas, \ie~sky, trees in background and shadowed areas difficult to interpret, compared to competitive unsupervised models~\cite{garg2016unsupervised,godard2017unsupervised,pilzer2018unsupervised}. Note that small details are better reconstructed by~\cite{godard2017unsupervised} but, overall, our estimations look smoother and have fewer large errors, as the train windshield in row seven.

\section{Conclusions}

We proposed a monocular depth estimation network which computes the inconsistencies
between input and cycle-reconstructed images and  exploit them to generate state-of-the-art depth predictions through a refinement network. We proved that distillation is an effective paradigm for depth estimation and improve the student network performance by transferring information from the refinement network. 
In future work we plan to further improve the distillation process by accounting for teacher and student confidence in the estimates. In this way we expect to better guide the learning process and correct more effectively prediction inconsistencies.

\section{Acknowledgement}

We want to thank the NVIDIA Corporation for the dona-
tion of the GPUs used in this project.

{\small
\bibliographystyle{ieee}

}

\clearpage
\appendix
\section*{Appendix}
We report some implementation details and report further experimental results. Note
that, qualitative results are also reported in the video file
attached to this document.

\section{Training Details}
In all our experiments, we use a learning rate equal to 1e-5 and batches composed of $8$ stereo image pairs. We employ the Adam optimizer, with momentum parameter and the weight decay set to $0.9$ and 2e-5, respectively. We used an NVIDIA Titan Xp with 12 GB of memory.

\noindent
\textbf{Analysis of Time Aspect.} The initial training of the \textit{half-cycle} for $10$ epochs takes approximately $2.7$ hours, and the \textit{backward-cycle} decoder for $5$ epochs takes $1$ hour. Joint training of the \textit{cycle} requires $3.5$ hours for $10$ epochs. Then, for the \textit{inconsistency-network} $2$ hours for $5$ epochs. Finally, the joint fine tuning with self-distillation for $10$ epochs requires about $6.5$ hours.

At testing time, depending on time constraints, the \textit{student} or \textit{teacher} network can be used. The \textit{student} takes $25$ ms while the \textit{teacher}, that requires propagation through the full network, $48.5$ ms.

\section{Experimental Results}
In this section, we present additional qualitative results, an ablation study of our proposed method on KITTI dataset~\cite{kitti}, and visualizations of the inconsistency.



\begin{figure*}[t!]
	\centering
	\includegraphics[width=\textwidth]{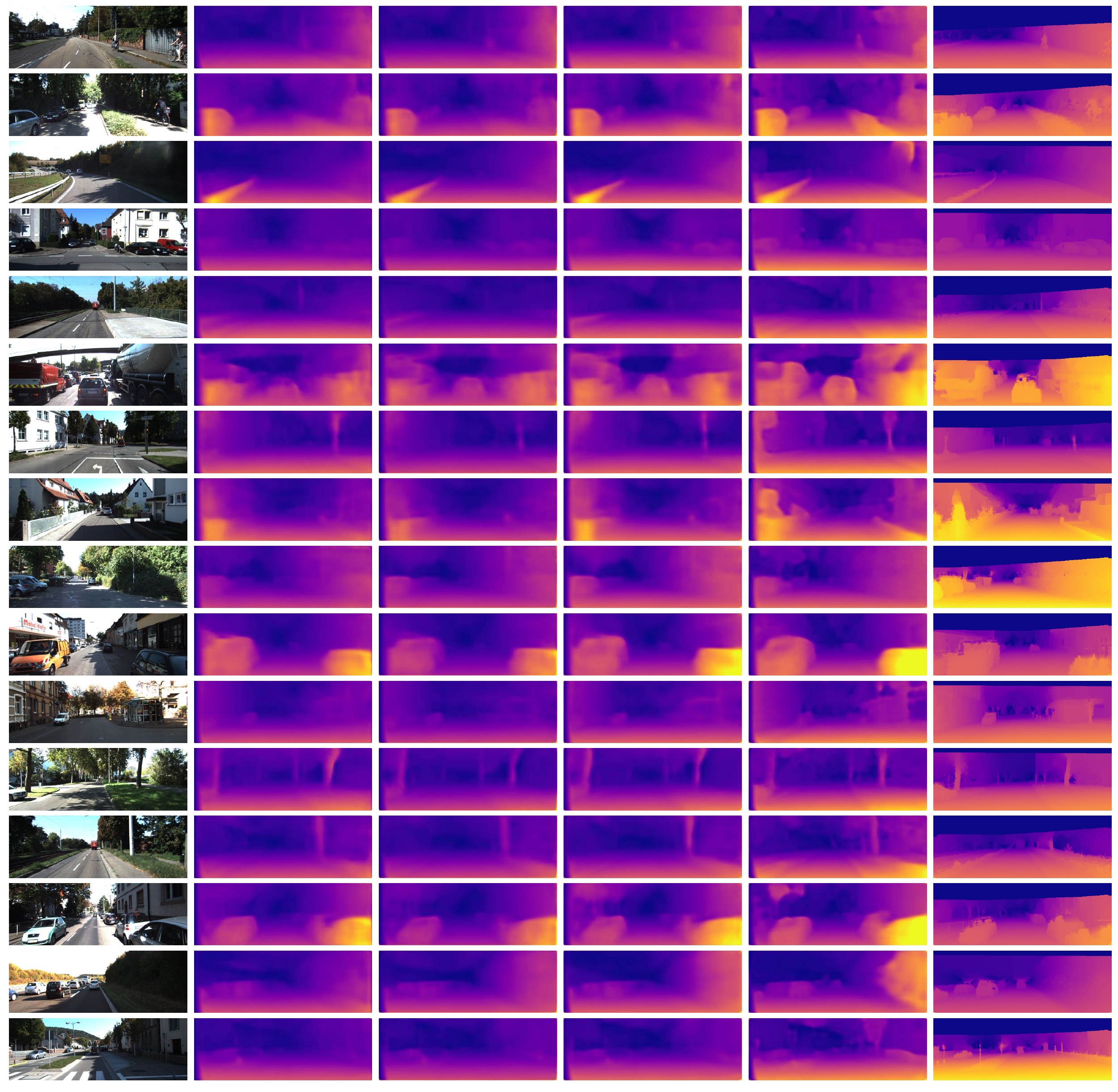}
	\put(-470,486){\scriptsize RGB Image}
	\put(-385,486){\scriptsize \textit{half-cycle}}
	\put(-295,486){\scriptsize \textit{cycle}}
	\put(-220,486){\scriptsize \textit{student}}
	\put(-140,486){\scriptsize \textit{teacher}}
	\put(-65,486){\scriptsize GT Depth Map}
\caption{Qualitative visualization of the ablation study on the KITTI test split proposed by Eigen~\etal~\cite{eigen2015predicting}.}
\label{fig:ablation_kitti}
\end{figure*}

In Fig.~\ref{fig:ablation_kitti}, we report a qualitative ablation study on the KITTI dataset. These results are consistent with the qualitative ablation study on Cityscapes and with the quantitative ablation on KITTI both reported in the main paper. Indeed, we first observe that our \textit{teacher} network estimates better the scene details,~\eg rows 1,3,4,6 and 8 where the image contains many trees and cars. For instance, in the first row, the depth of bicycle is not correctly estimated by our \emph{half-cycle}. The image in row 4 is a particularly interesting example since the image is challenging due to the presence of many vehicles. Again, we observe that our inconsistency-aware network (referred to as \emph{teacher}) predicts better depth maps.

\begin{figure*}[t!]
	\centering
	\includegraphics[width=\textwidth]{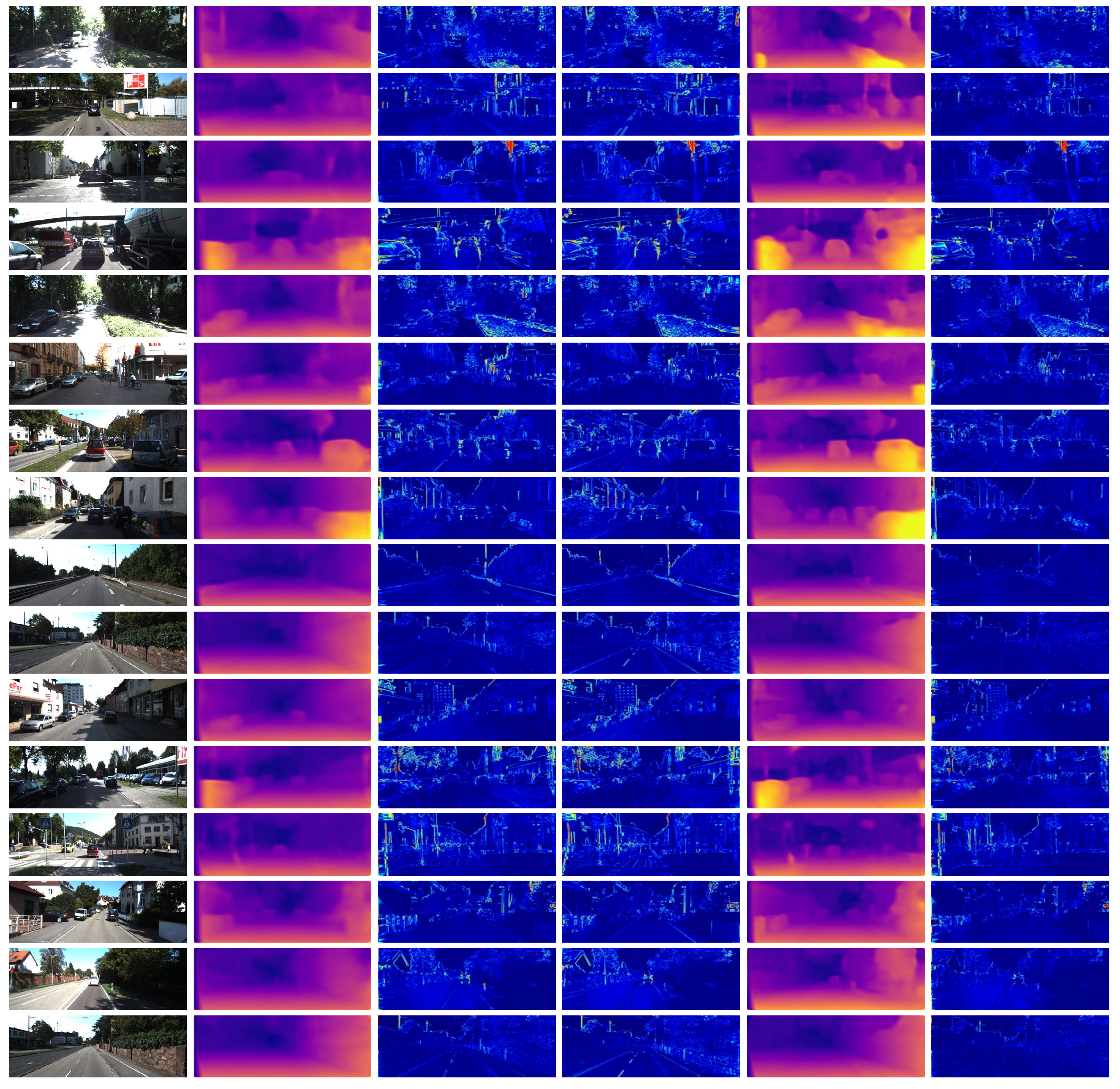}
	\put(-470,487){\scriptsize RGB Image}
	\put(-380,487){\scriptsize \textit{student}}
	\put(-310,487){\scriptsize \textit{student error}}
	\put(-225,487){\scriptsize \textit{inconsistency}}
	\put(-135,487){\scriptsize \textit{teacher}}
	\put(-65,487){\scriptsize \textit{teacher error}}
\caption{Qualitative comparison of \textit{student} and \textit{teacher} estimations, with cycle-inconsistencies and errors of the \textit{teacher} and \textit{student} on the KITTI test split proposed by Eigen~\etal~\cite{eigen2015predicting}.}
\label{fig:inconsistency_kitti}
\end{figure*}
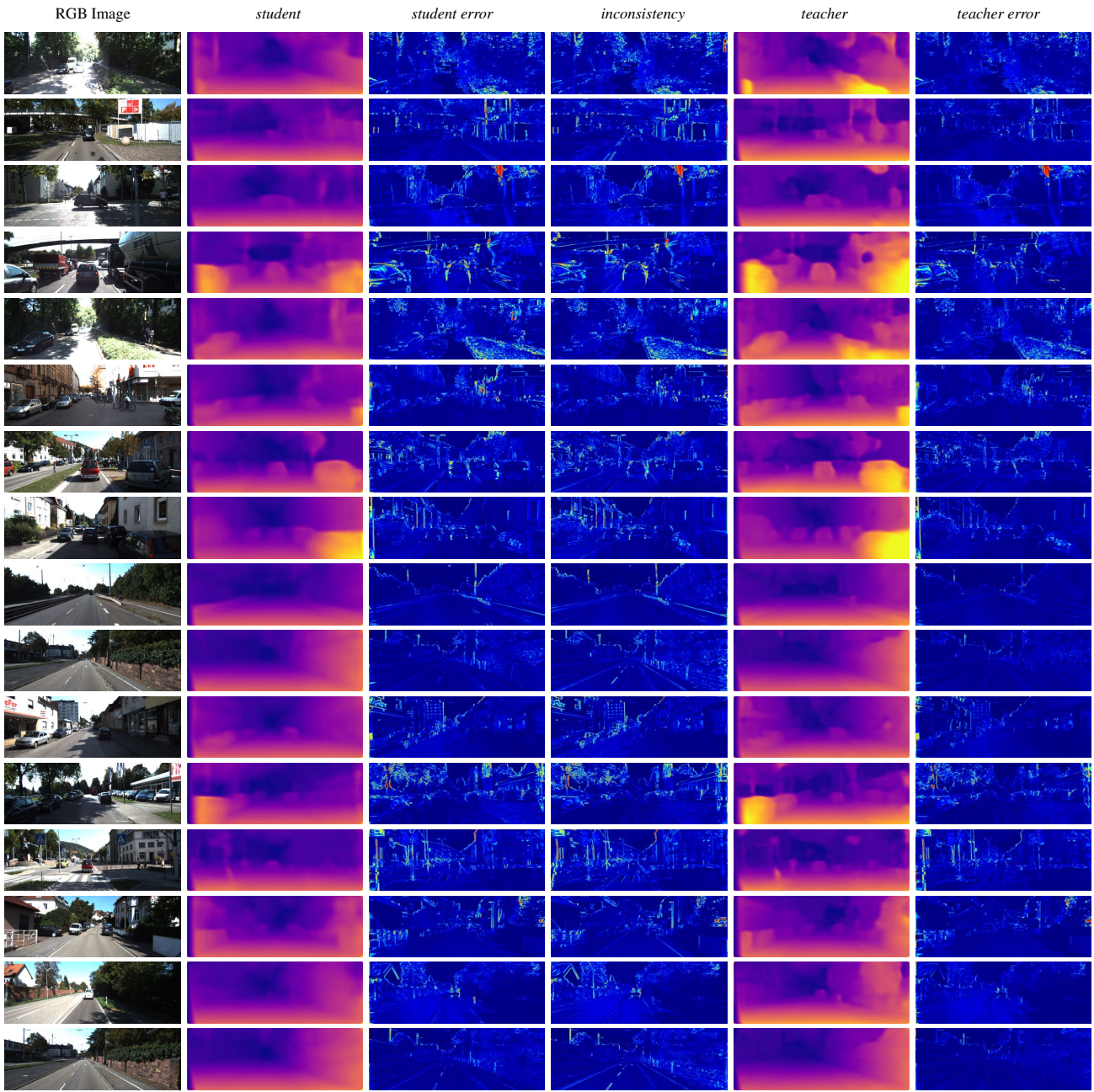

In order to further analyze the performance of our model, in Fig. \ref{fig:inconsistency_kitti}, we compare the inconsistency tensor, estimated by the \textit{cycle} network, with the reconstruction errors of the \textit{student} and \textit{teacher} networks. First, we observe that the inconsistency tensors, column $4$, are really similar to the reconstruction errors of the \textit{student}, column $3$. It shows that our cycle approach is able to estimate correctly the location of the errors in the student predictions. Second, most of the errors are located on the object edges. It confirms that, the cycle inconsistency can provide information about edge location. Third, comparing the reconstruction errors of the \textit{student}, column $3$, with the \textit{teacher}'s reconstruction errors, column $6$, we observe that the \emph{teacher}'s error maps contain much fewer large errors. For instance, in row 4, the \emph{student} network generates large errors on the edges of the car in the image center. Those errors are also visible on the inconsistency maps but are much smaller in the \emph{teacher} prediction. This better estimation of the car edges can be also observed by comparing the depth maps predicted by the \emph{student} and the \emph{teacher}.
In row 7 and in the last two rows, the student network generates errors on the dash lines on the road. These errors are also visible in the inconsistency tensors but are substantially reduced in the \emph{teacher} predictions. These examples clearly illustrate the benefit of our inconsistency-aware network.
Finally, in rows 1,2,3,5,9,10,12 and 15, we note that the \emph{student} generates many errors when the input image contains trees. The \emph{teacher} predictions are consistently better in the image regions containing trees.


\end{document}